\documentclass[letterpaper, 10 pt, journal, twoside]{IEEEtran}

\usepackage{cite}
\usepackage[pdftex]{graphicx}
\graphicspath{{./figures/}}
\DeclareGraphicsExtensions{.pdf,.jpeg,.png}
\usepackage[usenames,dvipsnames]{xcolor}

\usepackage[caption=false,font=footnotesize]{subfig}

\usepackage{amsmath,amssymb}
\usepackage{physics}
\usepackage{multirow}
\usepackage{amsthm}

\theoremstyle{plain}
\newtheorem{theorem}{Theorem}

\newtheorem{lemma}[theorem]{Lemma}
\newtheorem{proposition}{Proposition}

\theoremstyle{remark}

\theoremstyle{definition}
\newtheorem{definition}{Definition}

\usepackage{algorithm,algorithmic}

\usepackage{enumerate}

\usepackage{hyperref}
\hypersetup{
  colorlinks=true,
  linkcolor=black,
  urlcolor=black,
  citecolor=black
}

\usepackage{tikz}
\usepackage{textcomp}
\usepackage{lipsum}

\newcommand\copyrighttext{%
  \footnotesize \textcopyright 2025 IEEE.\@  Personal use of this material is permitted.  Permission from IEEE must be obtained for all other uses, in any current or future media, including reprinting/republishing this material for advertising or promotional purposes, creating new collective works, for resale or redistribution to servers or lists, or reuse of any copyrighted component of this work in other works.}
\newcommand\copyrightnotice{%
\begin{tikzpicture}[remember picture,overlay]
\node[anchor=south,yshift=10pt] at (current page.south) {\fbox{\parbox{\dimexpr\textwidth-\fboxsep-\fboxrule\relax}{\copyrighttext}}};
\end{tikzpicture}%
}

\begin{document}

\markboth{IEEE Robotics and Automation Letters. Preprint Version. July, 2025}
{Dry \MakeLowercase{\textit{et al.}}: ZORMS-LfD} 

\author{Olivia Dry$^{1}$, Timothy L. Molloy$^{1}$, Wanxin Jin$^{2}$, and Iman Shames$^{1}$%
\thanks{Manuscript received: March, 25, 2025; Revised June, 20, 2025; Accepted July, 15, 2025.}
\thanks{This paper was recommended for publication by Editor Lucia Pallottino upon evaluation of the Associate Editor and Reviewers' comments. 
This work was supported by the Australian Research Council under the Discovery Project DP250101763.} 
\thanks{$^{1}$O.\ Dry, T.\ L.\ Molloy, and I.\ Shames are with the CIICADA Lab, School of Engineering, The Australian National University (ANU), Canberra, ACT, Australia
        {\tt\footnotesize \{olivia.dry,timothy.molloy,iman.shames\}@anu.edu.au}}
\thanks{$^{2} $W.\ Jin is with the School for Engineering of Matter, Transport, and Energy, Arizona State University (ASU), Tempe, AZ, USA
        {\tt\footnotesize wanxin.jin@asu.edu}}
\thanks{Digital Object Identifier (DOI): see top of this page.}
}

\title{ZORMS-LfD: Learning from Demonstrations with Zeroth-Order Random Matrix Search}


\maketitle
\copyrightnotice

\begin{abstract}
We propose Zeroth-Order Random Matrix Search for Learning from Demonstrations (ZORMS-LfD).
ZORMS-LfD enables the costs, constraints, and dynamics of constrained optimal control problems, in both continuous and discrete time, to be learned from expert demonstrations without requiring smoothness of the learning-loss landscape.
In contrast, existing state-of-the-art first-order methods require the existence and computation of gradients of the costs, constraints, dynamics, and learning loss with respect to states, controls and/or parameters.
Most existing methods are also tailored to discrete time, with constrained problems in continuous time receiving only cursory attention.
We demonstrate that ZORMS-LfD matches or surpasses the performance of state-of-the-art methods in terms of both learning loss and compute time across a variety of benchmark problems.
On unconstrained continuous-time benchmark problems, ZORMS-LfD achieves similar loss performance to state-of-the-art first-order methods with an over $80$\% reduction in compute time.
On constrained continuous-time benchmark problems where there is no specialized state-of-the-art method, ZORMS-LfD is shown to outperform the commonly used gradient-free Nelder-Mead optimization method.
We illustrate the practicality of ZORMS-LfD on a human motion dataset, and derive complexity bounds for it on problems with Lipschitz continuous (but potentially nondifferentiable) loss. 
\end{abstract}


\section{Introduction}

\IEEEPARstart{W}{hen} designing robots for particular tasks, it is often helpful to observe demonstrations performed by experts.
To date, human demonstrations of walking, jumping, and reaching have informed the design of humanoid robots~\cite{locomotion_mombaur,Dana-Kulic-Motion,jumping,rebula2019robustness}; human-driver pathing and interaction data has inspired the design of technologies for autonomous vehicles~\cite{neumeyer2021general}; and, human-pilot demonstrations have been used to learn controllers for autonomous helicopter aerobatics~\cite{Abbeel2010}.
Techniques for learning from demonstrations (LfD) have been developed based on inverse optimal control (IOC) (or inverse reinforcement learning) approaches that learn the parameters of optimal control problems (e.g., costs, constraints, and dynamics) such that their solutions reproduce demonstration trajectories.
A key challenge of IOC is minimizing loss functions that penalize differences between optimal control solutions and demonstration trajectories, since these losses are often nonconvex, nonsmooth, and nondifferentiable.
We seek to overcome this challenge by introducing Zeroth-Order Random Matrix Search for LfD (ZORMS-LfD).

Despite the lack of regularity of the loss in many IOC problems, recent IOC approaches such as DiffMPC~\cite{DiffMPC}, PDP~\cite{PDP_framework} and its extensions, Safe-PDP~\cite{SafePDP} and CPDP~\cite{CPDP}, and IDOC~\cite{COC_problems}, are grounded in first-order (gradient) optimization methods.
These approaches establish the (local) existence and computability of gradients of the loss (and hence optimal control solutions), with respect to optimal control problem parameters via implicit function theorems or approximations. 
Such methods therefore lack theoretical convergence guarantees and complexity bounds, and can fail when learning constrained optimal control problems where constraints render the loss nonsmooth~\cite{SafePDP,COC_problems}.

Additionally, techniques for IOC in continuous time are far less developed than their discrete-time counterparts (cf.\ \cite{DiffMPC,PDP_framework,SafePDP,COC_problems,CPDP,Molloy2022}).
Indeed, CPDP~\cite{CPDP} appears to be the only state-of-the-art method of IOC in continuous time.
It is, however, only applicable when the optimal control problem to be learned is unconstrained, which is restrictive in many robotics settings.
IOC in continuous time with constraints is thus limited to basic approaches that employ general purpose optimization methods and lacking in guarantees (cf.\ \cite{locomotion_mombaur,Molloy2022,CPDP}).

The key contribution of this paper is ZORMS-LfD, a method of IOC for LfD that avoids imposing smoothness and differentiability conditions on the loss, and the costs, constraints, and dynamics of the optimal control problem to be learned.
ZORMS-LfD avoids these conditions by approximating the (potentially nonsmooth) loss and its derivatives using a zeroth-order random matrix oracle.
With appropriate oracles, ZORMS-LfD is applicable to \emph{all} IOC settings, both discrete- and continuous-time, and with and without constraints.
It also enjoys complexity bounds that act as convergence guarantees and guide hyperparameter selection.

This paper is structured as follows: Section~\ref{sec:relatedWork} presents related work; Section~\ref{sec:notation} describes our notation, Section~\ref{sec:problem} formulates IOC for LfD;\@ Section~\ref{sec:zorms} introduces ZORMS-LfD; Section~\ref{sec:experiments} reports the experimental setup; Section~\ref{sec:results} presents results; and, Section~\ref{sec:conclusion} offers conclusions.



\section{Related Work}\label{sec:relatedWork}
IOC approaches principally involve either residual minimization or bilevel optimization~\cite{Molloy2022}.
Residual minimization methods minimize residual functions that quantify the extent to which demonstrations violate optimality conditions satisfied by solutions to optimal control problems.
Residual functions have been based on Karush-Kuhn-Tucker (KKT) conditions~\cite{imputing,inverseKKT,convex_loco,jumping} and Pontryagin's principle~\cite{pontry} in discrete time, and Pontryagin's principle~\cite{cont_time,Molloy2022}, Euler-Lagrange equations~\cite{diff_flat}, and Hamilton-Jacobi-Bellman conditions~\cite{IOC_polynom} in continuous time.
Whilst residual minimization methods are computationally efficient, they handle noisy partial demonstrations and constraints poorly~\cite{robust_approach, Molloy2022}.

Bilevel methods find optimal control parameters by minimizing loss functions that penalize differences between demonstration trajectories and solutions of the optimal control problem~\cite{Molloy2022}.
They have found widespread use due to the flexibility of specifying meaningful  loss functions that capture the essential IOC aim of finding an optimal control problem whose solutions reproduce demonstration trajectories~\cite{locomotion_mombaur}.
Bilevel methods optimize the (upper-level) loss function by solving the (lower-level) optimal control problem with different candidate parameters.
A core challenge with bilevel methods is thus finding efficient procedures for the (upper-level) loss optimization.
Naive implementations using optimization methods that compute numerical gradients lead to the need to solve numerous optimal control problems.

Early bilevel approaches employ gradient-free search methods similar to the Nelder-Mead algorithm~\cite{locomotion_mombaur}.
Recent approaches such as DiffMPC~\cite{DiffMPC}, PDP~\cite{PDP_framework}, Safe-PDP~\cite{SafePDP}, CPDP~\cite{CPDP}, and IDOC~\cite{COC_problems} have employed first-order methods using analytical gradients computed via implicit function theorems or (local) approximations under certain smoothness assumptions on the loss, costs and dynamics, which are difficult to verify and can fail to hold.
Such approaches also lack guarantees such as bounds on their complexity and convergence, and all except CPDP~\cite{CPDP} are tailored to discrete time (with CPDP only handling unconstrained continuous-time problems).
We avoid smoothness assumptions and handle constrained discrete- and continuous-time problems by exploiting zeroth-order optimization, which uses random oracles instead of gradients, and has found success in nonsmooth problems in planning through contact~\cite{suh2022bundled}, and optimal control and reinforcement learning~\cite{suh2022differentiable,le2024leveraging,zorms}.


{\color{black}
\section{Notation}\label{sec:notation}
The set of natural numbers up to $N$ is $\mathbb{N}_N \triangleq \{0, 1,\ldots, N\}$.
The set of real $n$-d vectors is $\mathbb{R}^{n}$, with $\mathbb{R}^1 = \mathbb{R}$, and $\mathbb{R}^{n\times m}$ is the set of $n \times m$ real matrices. 
The set of real $n \times n$ symmetric matrices is $\mathbb{S}^{n}$, the set of real $n \times n$ symmetric positive semidefinite matrices is $\mathbb{S}^{n}_+$, and $\mathbb{S}^{n}_{++}$ is the set of real $n \times n$ symmetric positive definite matrices. 
For a matrix $A$, its transpose is $A^\top$, ${[A]}_{ij}$ denotes the element in its $i$-th row and $j$-th column, and $\norm{A}_F$ is its Frobenius norm.
Given an ordered collection of square matrices $(A_1,\ldots,A_N)$, $\mathrm{blkdiag}(A_1,\ldots,A_N)$ is the block diagonal matrix with the matrices $(A_1,\ldots,A_N)$ in order on its diagonal.
The $\ell_2$-norm is $\norm{\cdot}_2$, and the $n \times n$ identity matrix is $I_n$.
Finally, $x \sim \mathcal{N}(\mu,\sigma^2)$ denotes that $x \in \mathbb{R}$ is normally distributed with mean $\mu \in \mathbb{R}$ and variance $\sigma^2 > 0$.%
}

\section{Problem Formulation}\label{sec:problem}


We focus initially on learning parameters of continuous-time optimal control problems from noisy partial measurements of demonstrations since the existing state-of-the-art~\cite{CPDP} is restricted to problems without constraints.
To this end, consider the continuous-time optimal control problem
\begin{align}
    \label{eq:OCP-CT}
    \begin{split}
        \min_{x,u} \quad & h(x(T);\theta) + \int_{0}^{T} \ell(x(t), u(t);\theta) \; \mathrm{d} t \\
        \text{s.t.} \quad & \dot{x}(t) = f(x(t), u(t);\theta), \quad x(0) = x_0 \\
        & c_t(x(t), u(t); \theta) \leq 0, \quad \forall t\in[0,T) \\
        & \bar{c}_t(x(t), u(t); \theta) = 0, \quad \forall t\in[0,T) \\
        & c_T(x(T); \theta) \leq 0, \quad \bar{c}_T(x(T); \theta) = 0
    \end{split}
\end{align}
for (continuous) time $t \in [0,T]$ where $0 < T < \infty$ is a finite horizon; $x(t) \in \mathbb{R}^n$ are the states; $u(t) \in \mathbb{R}^m$ are the controls; $f:\mathbb{R}^n\times\mathbb{R}^m\rightarrow\mathbb{R}^n$ is the system dynamics; the stage/running cost $\ell:\mathbb{R}^n\times\mathbb{R}^m\rightarrow\mathbb{R}$ and the terminal cost $h:\mathbb{R}^n\rightarrow\mathbb{R}$; and the problem may involve both path and terminal inequality constraints $c_t:\mathbb{R}^n\times\mathbb{R}^m\rightarrow\mathbb{R}^{q_t}$ and $c_T:\mathbb{R}^n\rightarrow\mathbb{R}^{q_T}$, as well as path and terminal equality constraints $\bar{c}_t:\mathbb{R}^n\times\mathbb{R}^m\rightarrow\mathbb{R}^{s_t}$ and $\bar{c}_T:\mathbb{R}^n\rightarrow\mathbb{R}^{s_T}$.

The dynamics $f$, costs $\ell$, and constraints $c_t$ and $\bar{c}_t$ of the optimal control problem~\eqref{eq:OCP-CT} are parameterized by a collection of symmetric matrices $( \theta_1, \theta_2, \ldots, \theta_\rho)$ where $\theta_i \in \Theta_i$ and $\Theta_i \subseteq \mathbb{S}^{p_i}$ for $i = 1, 2, \ldots, \rho$ are subsets of symmetric matrices (e.g., $\Theta_i$ could be the set of positive definite or positive semidefinite matrices).
We collect the matrices $( \theta_1, \theta_2, \ldots, \theta_\rho)$ in the (symmetric) block-diagonal matrix $\theta \triangleq \mathrm{blkdiag}(\theta_1, \theta_2, \ldots, \theta_\rho) \in \Theta$ from the set
\begin{align*}	
	\Theta
	&\triangleq \{ \mathrm{blkdiag}(\theta_1, \theta_2, \ldots, \theta_\rho) : \theta_i \in \Theta_i \text{ for } i = 1, 2, \ldots, \rho\}.
\end{align*}
Here, $\Theta$ is a closed convex subset of $\mathbb{S}^p$ for $p \triangleq \sum_{i=1}^{\rho} p_i$.\footnote{Despite most existing IOC approaches employing vector parameterizations, this matrix parameterization is not restrictive and will prove useful in developing ZORMS-LfD and its complexity bounds. Indeed, given a parameter vector from an existing work, a corresponding symmetric matrix parameter can be found by simply constructing a diagonal matrix with the elements of the parameter vector on its diagonal.}

In (forward) optimal control, the aim is to find optimal states $x(t;\theta)$ and controls $u(t;\theta)$ solving~\eqref{eq:OCP-CT} given parameters $\theta$.
However, in IOC for LfD we observe expert demonstrations in the form of optimal states and controls solving~\eqref{eq:OCP-CT} with an unknown $\theta = \theta^*$ through measurements
\begin{equation}
    \label{eq:measurements-CT}
    y(t_i) = g(x(t_i;\theta^*), u(t_i;\theta^*)) + w(t_i)
\end{equation}
at sample times $0\leq t_1<\cdots<t_\tau\leq T$ where $i = 1, \ldots, \tau$, $g:\mathbb{R}^n\times\mathbb{R}^m\rightarrow \mathbb{R}^q$ is a (partial) measurement function, and $w :[0, T] \rightarrow \mathbb{R}^q$ is a (potentially stochastic) noise process.
Given the measurements $\{y(t_1),\ldots,y(t_\tau)\}$, we aim to compute $\hat{\theta} \triangleq \mathrm{blkdiag} ( \hat{\theta}_1, \hat{\theta}_2, \ldots, \hat{\theta}_\rho)$ that approximates $\theta^*$ in~\eqref{eq:OCP-CT} by solving the bilevel loss-minimization problem
\begin{align}
    \label{eq:prob-CT}
    \begin{split}
        \underset{\theta \in \Theta}{\text{min}} \quad & \mathcal{L}(\theta) \triangleq \sum_{i=1}^{\tau} L\left(y(t_i),g(x(t_i;\theta), u(t_i;\theta))\right) \\
        \text{s.t.} \quad & x(\cdot; \theta), u(\cdot; \theta) \text{ solving~\eqref{eq:OCP-CT} for $\theta$} 
    \end{split}
\end{align}
where $\mathcal{L} : \Theta \rightarrow \mathbb{R}$ is the \emph{loss} function, and $L$ is a scalar function that quantifies the difference between the (measured) demonstration states and controls, and the solution of~\eqref{eq:OCP-CT} for candidate parameters $\theta$, at measurement instance $i$, with the squared Euclidean norm $L\left(y(t_i),g(x(t_i;\theta), u(t_i;\theta))\right) = \norm{y(t_i) - g(x(t_i;\theta),u(t_i;\theta))}_2^2$ being a common choice.

The bilevel problem~\eqref{eq:prob-CT} is difficult to solve with first-order methods since gradients of the loss $\mathcal{L}$ can fail to exist, or become hard to compute, when it is nonsmooth (e.g., when constraints in~\eqref{eq:OCP-CT} are active).
To solve~\eqref{eq:prob-CT}, we propose \emph{ZORMS-LfD}, a method that uses zeroth-order oracles in place of gradients and has theoretical performance guarantees.


\section{ZORMS-LfD}\label{sec:zorms}

In this section, we present ZORMS-LfD and develop its performance guarantees via complexity bounds.

\subsection{ZORMS-LfD Algorithm}
To find the (matrix) parameters $\hat{\theta} \in \Theta$ solving the loss-minimization problem~\eqref{eq:prob-CT}, ZORMS-LfD involves iterating
\begin{equation}
    \label{eq:iterates}
    \hat{\theta}_{k+1} 
    = \mathbb{P}_{\Theta}\left[\hat{\theta}_k - \alpha_k\mathcal{O}_\mu(\hat{\theta}_k, M^U_k)\right]
\end{equation}
for $k \in \mathbb{N}_N$ given an initial $\hat{\theta}_0 \in \Theta$ where $N > 0$ is the desired number of iterations, $\alpha_k > 0$ are chosen step sizes, and $\mathbb{P}_{\Theta}$ is the Euclidean projection onto $\Theta$.
Here, $\mathcal{O}_\mu$ is the zeroth-order random matrix oracle defined as
\begin{equation}
    \label{eq:zo-oracle}
    \mathcal{O}_\mu(\hat{\theta}_k, M_k^U)
    \triangleq \left[ \mathcal{L}(\hat{\theta}_k + \mu M_k^U)-\mathcal{L}(\hat{\theta}_k) \right] \mu^{-1} M_k^U
\end{equation}
where $\mu > 0$ is the oracle's desired precision, and $M_k^U \triangleq \mathrm{blkdiag}(U_k^1,U_k^2, \ldots, U_k^\rho)$ for $k \in \mathbb{N}_N$ are symmetric block-diagonal matrices with $U_k^i \in \mathbb{G}^{p_i}$ for $i=1,2,\ldots,\rho$ from the Gaussian Orthogonal Ensemble $\mathbb{G}^{p_i}$, defined as follows.

\begin{definition}[Gaussian Orthogonal Ensemble~\cite{Anderson_Guionnet_Zeitouni_2009}]\label{def:GOE}
    The Gaussian Orthogonal Ensemble (GOE), denoted $\mathbb{G}^{p}$, is the set of random symmetric matrices $U\in\mathbb{S}^{p}$ with entries such that ${[U]}_{ii}\sim \mathcal{N}(0,1)$ and ${[U]}_{ij}\sim \mathcal{N}(0,\frac{1}{2})$ are independent for $i<j$, and ${[U]}_{ij} = {[U]}_{ji}$ where $1 \leq i,j \leq p$.
\end{definition}

After iterating~\eqref{eq:iterates} the desired number of times $N$, we select the final (learned) parameters $\hat{\theta}$ from $\{\hat{\theta}_0, \hat{\theta}_1, \ldots, \hat{\theta}_N\}$ as those that achieve the smallest value of the loss, namely,
\begin{align*}
	\hat{\theta} \in  \arg \min_\theta \left\{\mathcal{L}(\theta) : \theta \in \{\hat{\theta}_0, \hat{\theta}_1, \ldots, \hat{\theta}_N\}\right\}.
\end{align*}
ZORMS-LfD is summarized in Algorithm~\ref{alg:ZO-RMS}.

The key iteration~\eqref{eq:iterates} of ZORMS-LfD resembles that of first-order optimization methods with the oracle~\eqref{eq:zo-oracle} instead of analytical or numerical gradients.
{\color{black}
Computing~\eqref{eq:zo-oracle} involves solving the (forward) optimal control problem~\eqref{eq:OCP-CT} twice, once with the parameters $\theta = \hat{\theta}_k$ for $\mathcal{L}(\hat{\theta}_k)$, and once with the parameters $\theta = \hat{\theta}_k + \mu M_k^U$ for $\mathcal{L}(\hat{\theta}_k + \mu M_k^U)$.
Importantly, unlike existing gradient-based algorithms~\cite{CPDP,SafePDP,COC_problems}, these optimal control problems are independent, enabling their solution in a (embarrassingly) parallel manner.

The oracle~\eqref{eq:zo-oracle} can be interpreted as constructing a smoothed estimate of the loss function $\mathcal{L}$ by (iteratively) examining its directional derivatives in the random directions $M_k^U$, with $\mu$ controlling the degree of smoothing (see \cite{vec_oracle,zorms} for details and discussions of Gaussian smoothing).
ZORMS-LfD thus handles nonsmooth loss functions $\mathcal{L}$.
Hence, it entirely avoids the differentiability conditions required by the existing state-of-the-art method of IOC in continuous time~\cite{CPDP}, and enables IOC in continuous time via~\eqref{eq:prob-CT} when the underlying optimal control problem~\eqref{eq:OCP-CT} is subject to constraints.
The theoretical significance of ZORMS-LfD is its complexity bounds that provide both convergence guarantees and a guide for choosing appropriate values of its hyperparameters $N$, $\mu$, and $\alpha_k$.
}

\begin{algorithm}[t!]
    \caption{ZORMS-LfD}\label{alg:ZO-RMS}
    \begin{algorithmic}
        \STATE{} {\bfseries Input:} Measurements $\{y(t_1),\ldots,y(t_\tau)\}$, horizon $T$.
        \STATE{} Choose $\hat{\theta}_0 \in \Theta$, $\mu>0$, $\{\alpha_k\}>0$, and $N > 0$.
        \FOR{$k \in \mathbb{N}_N$}
        	\STATE{} Generate $U_k^i \in \mathbb{G}^{p_i}$ for $i=1,2,\ldots,\rho$.
            \STATE{} Construct $M_k^U = \mathrm{blkdiag}(U_k^1,U_k^2, \ldots, U_k^\rho)$.
	        \STATE{} Solve~\eqref{eq:OCP-CT} with $\theta = \hat{\theta}_k$ to compute ${\{x(t_i;\hat{\theta}_k)\}}_{i=1}^{\tau}$\\ \hfill and ${\{u(t_i;\hat{\theta}_k)\}}_{i=1}^{\tau}$.
	        \STATE{} Evaluate $\mathcal{L}(\hat{\theta}_k)$.
	        \STATE{} Solve~\eqref{eq:OCP-CT} with $\theta = \hat{\theta}_k + \mu M_k^U$ to compute\\ \hfill ${\{x(t_i;\hat{\theta}_k+\mu M^U_k)\}}_{i=1}^{\tau}$ and ${\{u(t_i;\hat{\theta}_k+\mu M^U_k)\}}_{i=1}^{\tau}$.
	        \STATE{} Evaluate $\mathcal{L}(\hat{\theta}_k + \mu M_k^U)$.
            \STATE{} Evaluate $\mathcal{O}_\mu\gets \frac{1}{\mu}(\mathcal{L}(\hat{\theta}_k + \mu M^U_k)-\mathcal{L}(\hat{\theta}_k))M^U_k$.
            \STATE{} Update $\hat{\theta}_{k+1}\gets \mathbb{P}_{\Theta}\left[\hat{\theta}_k - \alpha_k\mathcal{O}_\mu(\hat{\theta}_k, M^U_k)\right]$.
        \ENDFOR{}
        \RETURN{} $\hat{\theta} \in  \arg \min_\theta \{\mathcal{L}(\theta) : \theta \in \{\hat{\theta}_0, \hat{\theta}_1, \ldots, \hat{\theta}_N\}\}$
    \end{algorithmic}
\end{algorithm}

\subsection{Complexity Bounds and Hyperparameter Selection}



To establish complexity bounds for ZORMS-LfD, we exploit properties of the GOE (cf.\ Definition~\ref{def:GOE}).
Specifically, note that the GOE implies the probability distribution
\begin{equation}
    \label{eq:GOE-dist}
    P(\mathrm{d}U) = \kappa^{-1} e^{-\frac{1}{2}\norm{U}^2_F} \; \mathrm{d} U
\end{equation}
on $U \in \mathbb{S}^p \cong \mathbb{R}^{p(p+1)/2}$ where $\mathrm{d}U$ is the Lebesgue measure on $\mathbb{S}^p \cong \mathbb{R}^{p(p+1)/2}$ and $\kappa$ is a normalizing constant so that~\eqref{eq:GOE-dist} integrates to $1$ over $\mathbb{S}^p$.
Let $E[\cdot]$ denote the expectation operator corresponding to~\eqref{eq:GOE-dist}.
This probability distribution and expectation imply a corresponding probability distribution and expectation for (random) block-diagonal matrices $M^U \triangleq \mathrm{blkdiag}(U^1,U^2,\ldots,U^\rho)$ of the form used in the oracle~\eqref{eq:zo-oracle} where $U^i \in \mathbb{G}^{p_i}$ for $i = 1, 2, \ldots, \rho$.
The following lemma expresses and bounds moments of this distribution for use in deriving complexity bounds for ZORMS-LfD.
\begin{lemma}\label{lemma:moments}
	Consider $M^U =
	\mathrm{blkdiag}(U^1,U^2,\ldots,U^\rho)$ with $U^i \in
	\mathbb{G}^{p_i}$, $p_i \geq 1$ for $i = 1, 2, \ldots, \rho$.
	Define $m_1 \triangleq E[\|M^U\|_F]$, $m_2 \triangleq
	E[\|M^U\|_F^2]$, and $m_4 \triangleq E[\|M^U\|_F^4]$. Then, 
	$
        m_1 \leq \sqrt{\frac{1}{2} \sum_{i=1}^{\rho} (p_i^2 + p_i)}$,
        $m_2 = \frac{1}{2} \sum_{i=1}^{\rho} (p_i^2 + p_i)
	$, 
	and
	$m_4 = \frac{1}{4} \sum_{i = 1}^{\rho} \left( p_i^4 + 2p_i^3 + 5p_i^2 + 4p_i \right) + \frac{1}{2} \sum_{i,j = 1, i \neq j}^\rho (p_i^2 + p_i)(p_j^2 + p_j)$.
\end{lemma}
\begin{proof}
    Since $M^U$ is block-diagonal, its Frobenius norm is $\|M^U\|^2_F = \sum_{i=1}^\rho \|U^i\|_F^2 $.
    Moreover, $U^i$ and $U^j$ are independent random matrices so $m_2 = E[\sum_{i=1}^\rho \|U^i\|_F^2] = \sum_{i=1}^\rho E[\|U^i\|_F^2]$.
    Further, $m_4 = E[{(\sum_{i=1}^\rho \|U^i\|_F^2)}^2] = \sum_{i=1}^\rho E[\|U^i\|_F^4] + 2\sum_{i,j = 1,i\neq j}^\rho E[\|U^i\|_F^2] E[\|U^j\|_F^2]$.
    From~\cite{moments}, we have that $E[\|U^i\|_F^2] = \frac{1}{2}(p_i^2+p_i)$ and $E[\|U^i\|_F^4] = \frac{1}{4}(p_i^4 + 2p_i^3 + 5p_i^2 + 4p_i)$.
    Finally, from~\cite{zorms}, we have that $m_1\leq \sqrt{m_2}$, which concludes the proof.
\end{proof}

Given Lemma~\ref{lemma:moments}, we establish our first complexity bound.

\begin{proposition}\label{proposition:convex}
    Suppose that the loss $\mathcal{L} : \Theta \rightarrow \mathbb{R}$ in~\eqref{eq:prob-CT} is convex and Lipschitz continuous in the sense that there exists a constant $\lambda(\mathcal{L})>0$ such that $|\mathcal{L}(X) - \mathcal{L}(Y)| \leq \lambda(\mathcal{L})\norm{X-Y}_F$ for all $X,Y\in \Theta$.
    Then for any $\epsilon>0$, ZORMS-LfD (Algorithm~\ref{alg:ZO-RMS}) yields $E [ \mathcal{L}(\hat{\theta})] - \mathcal{L}(\theta^*) \leq \epsilon$ if
    \begin{align*}
        \mu \leq \frac{\epsilon}{\lambda(\mathcal{L})\sqrt{4m_2}} \text{ and }
        \alpha_k = \frac{\bar{r}}{\lambda(\mathcal{L})\sqrt{m_4}\sqrt{k+1}}
    \end{align*}
    for $k\in \mathbb{N}_N$ with $N \geq 4m_4 \lambda^2(\mathcal{L})\bar{r}^2 \epsilon^{-2}$ and $\bar{r} \geq \|\hat{\theta}_0 - \theta^*\|_F$.
\end{proposition}
\begin{proof}
    The proof follows from those of~\cite[Th.~6]{vec_oracle} and~\cite[Th.~1, Cor.~1]{zorms} by substituting the appropriate expressions and bounds for the moments $m_1$, $m_2$, and $m_4$ established in Lemma~\ref{lemma:moments} for the matrices $M_k^U$.
\end{proof}


Proposition~\ref{proposition:convex} provides insight into the complexity of ZORMS-LfD and how to select its hyperparameters when the loss $\mathcal{L}$ in~\eqref{eq:prob-CT} is convex.
However, in most LfD problems, the loss $\mathcal{L}$ is likely to be nonconvex.
In the next proposition, we shall therefore develop a complexity bound for ZORMS-LfD for general (potentially nonconvex) loss functions $\mathcal{L}$.
To do so, let us define the Gaussian approximation of $\mathcal{L}$ as
    \begin{equation}
    	\label{eq:gaussianApprox}
        \mathcal{L}_\mu(\theta) \triangleq \frac{1}{\kappa} \int_{\mathbb{G}^p} \mathcal{L}(\theta + \mu M^U)e^{-\frac{1}{2}\norm{M^U}^2_F} \; \mathrm{d} M^U,
    \end{equation}
    for $\theta \in \Theta$, and define the \emph{gradient mapping} as~\cite{ghadimi2013,Liu2018}
\begin{equation*}
    G_\Theta (\theta, \mathcal{O}_\mu(\theta, M^U), \alpha) 
    \triangleq \frac{1}{\alpha}\left(\theta - \mathbb{P}_{\Theta}[\theta - \alpha \mathcal{O}_\mu (\theta, M^U)]\right).
\end{equation*}
Our main theoretical result follows.
    

\begin{proposition}\label{proposition:general}
    Suppose that the loss $\mathcal{L} : \Theta \rightarrow \mathbb{R}$ in~\eqref{eq:prob-CT} is Lipschitz continuous in the sense that there exists a constant $\lambda(\mathcal{L})>0$ such that $|\mathcal{L}(X) - \mathcal{L}(Y)| \leq \lambda(\mathcal{L})\norm{X-Y}_F$ for all $X,Y\in \Theta$.
    Then, for any $\varepsilon > 0$ and $\delta > 0$, if
    \begin{align*}
        \mu = \frac{\varepsilon}{\lambda(\mathcal{L})\sqrt{m_2}} \text{ and }
        \alpha_k = \alpha \triangleq {\left[\frac{\varepsilon\bar{r}}{(N+1)\lambda^3(\mathcal{L})m_2m_4}\right]}^{1/2}
    \end{align*}
    for all $k \in \mathbb{N}_N$ with $N \geq 2m_2m_4\lambda^5(\mathcal{L})\bar{r}\varepsilon^{-1}\delta^{-2}$ and $\bar{r} \geq \|\hat{\theta}_0 - \theta^*\|_F$, then $|\mathcal{L}_\mu(\theta) - \mathcal{L}(\theta)| < \varepsilon$ for all $\theta \in \Theta$, and ZORMS-LfD (Algorithm~\ref{alg:ZO-RMS}) achieves $E[\|G_{\Theta}(\hat{\theta}_D, \mathcal{O}_\mu(\hat{\theta}_D, M^U_D), \alpha)\|^2_F] \leq \delta + \lambda^2(\mathcal{L})m_4$ where $D \triangleq \arg \underset{k \in \mathbb{N}_N}{\min} \|G_{\Theta}(\hat{\theta}_k, \mathcal{O}_\mu(\hat{\theta}_k, M^U_k), \alpha)\|_F$.
\end{proposition}
\begin{proof}
	From~\cite[Sec.~3.2]{zorms} and~\cite[Th.~3]{zorms}, we have
    	$E[\norm{\mathcal{O}_\mu (\theta, M^U) - \nabla\mathcal{L}_\mu(\theta)}^2_F]
    	\leq E[\|\mathcal{O}_\mu (\theta, M^U)\|_F^2]
    	\leq \lambda^2(\mathcal{L}) m_4$
    for all $\theta \in \Theta$ where $\nabla\mathcal{L}_\mu(\theta)$ is the gradient of the Gaussian approximation~\eqref{eq:gaussianApprox}.
    The conditions of~\cite[Th.~2, Cor.~2]{zorms} are then met, and the assertion follows by applying the arguments in their proofs, substituting the expressions and bounds for $m_1$, $m_2$, and $m_4$ from Lemma~\ref{lemma:moments}.
\end{proof}



\begin{figure}[!t]
    \centering
    \subfloat[]{\includegraphics[width=0.47\columnwidth]{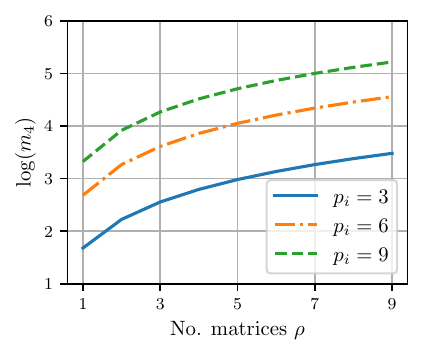}\label{fig:comp-bounds-dim}}
    \hfill
    \subfloat[]{\includegraphics[width=0.47\columnwidth]{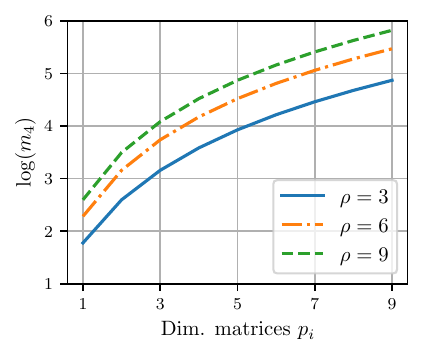}\label{fig:comp-bounds-num}}
    \hfill
    \subfloat[]{\includegraphics[width=0.54\columnwidth]{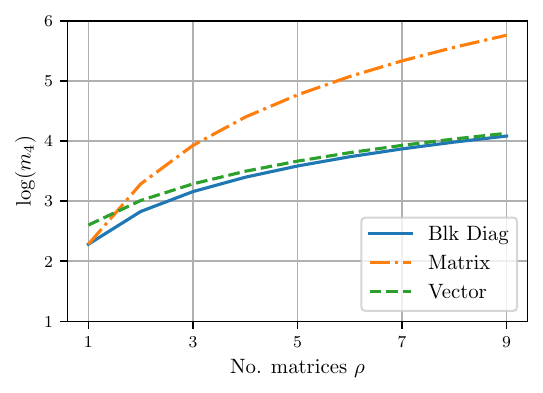}\label{fig:comp-bounds-comparison}}\hfill
    \caption{\color{black}Comparison of complexity bounds for different parameter-space $\Theta$ dimensions: (a) Set parameter matrix size with varying number of matrices, (b) Set number of matrices with varying dimension. (c) Shows the comparison with previous work, where the solid blue line depicts our block diagonal matrix bounds, the orange dashed line depicts the matrix bounds in~\cite{zorms}, and the green dashed depicts the vector bounds in~\cite{vec_oracle}.
    \vspace{-1.5em}
    }\label{fig:comp-bounds}
\end{figure}

Propositions~\ref{proposition:convex} and~\ref{proposition:general} represent the first theoretical convergence guarantees for IOC in continuous time with constraints (and indeed with noisy partial measurements).
Comparing Propositions~\ref{proposition:convex} and~\ref{proposition:general}, we see that it is possible to guarantee that ZORMS-LfD returns a parameter $\hat{\theta}$ such that the loss is arbitrarily small when the loss is convex.
However, in the general case of a nonconvex loss, the gradient mapping can be interpreted as the projected gradient offering a (feasible) update from $\theta$, and hence will be small when ZORMS-LfD is near convergence, providing a natural measure of stationarity.

{\color{black}The scalability of the (lower) bounds on the number of iterations ($N$) established in Propositions~\ref{proposition:convex} and~\ref{proposition:general} and the dimension of the parameter space $\Theta$ (which determines only $m_4$) is illustrated in Fig.~\ref{fig:comp-bounds}.
In both cases, the total dimension of the space $p$ is equal, but the structure of the block matrix differs.
We can see that the bound is dependent on both the total size of the parameter space $p$, and the structure of the space.}
{\color{black}To compare our complexity bounds to the bounds in~\cite{zorms} and~\cite{vec_oracle}, we can represent our block diagonal matrix $\theta$ as \color{black}a single symmetric matrix of dimension $p$, and a vector of the lower triangular elements of dimension $\sum_{i=1}^{\rho}\left(\frac{p_i(p_i + 1)}{2}\right)$, respectively.
The respective moments used in the bounds are:
\begin{align*}
    m_4^{[24]} &= \frac{1}{4}(p^4 + 2p^3 + 5p^2 + 4p) \\
    m_4^{[26]} &= {\left(\sum_{i=1}^{\rho}\left(\frac{p_i(p_i + 1)}{2}\right) + 4\right)}^2.
\end{align*}
Figure~\ref{fig:comp-bounds-comparison} depicts these bounds for increasing parameter dimensionality.
We can see that our bound is less conservative than those in~\cite{zorms} and~\cite{vec_oracle}, which follows from exploiting the block diagonal structure of the parameter space.
}

In both Propositions~\ref{proposition:convex} and~\ref{proposition:general}, we have assumed that the loss function is Lipschitz continuous, with the associated Lipschitz constant then appearing in expressions for the hyperparameters $N$, $\mu$, and $\alpha_k$.
We note that this continuity assumption is milder than an assumption of differentiability (or continuous differentiability) of the loss.
%
{\color{black} Additionally, while the selection of hyperparameters relies on the value of the Lipschitz constant, it is usually not feasible to explicitly determine it.
Instead, it is common in experimental optimization to estimate the Lipschitz constant from experimental data, and ensure that the condition is at least satisfied~\cite{bunin2017lipschitzconstantsexperimentaloptimization}.
Since the oracle uses the value of the loss at two different points, the Lipschitz constant can be estimated by selecting $\lambda(\mathcal{L}) \geq \max_{(\theta, \bar{\theta}) \in \Theta \times \Theta}(|\mathcal{L}(\theta) - \mathcal{L}(\bar{\theta})| \|\theta - \bar{\theta}\|_F^{-1})$.
Similarly, $\bar{r}$ can be estimated as the radius of the set $\Theta$ under the Frobenius norm.
}

Before demonstrating the practical performance of ZORMS-LfD, we shall illustrate its versatility by generalizing it to solving IOC for LfD in discrete time.

\subsection{ZORMS-LfD for IOC in Discrete Time}
Unlike existing state-of-the-art IOC methods, ZORMS-LfD is not specialized to the form of the underlying optimal control problem being learned, and thus can also be readily applied to IOC problems in discrete time.
To apply ZORMS-LfD in discrete time, the (forward) optimal control problems solved by the oracle $\mathcal{O}_\mu$ are simply replaced by discrete-time optimal control problems.
Specifically, in \emph{discrete-time} IOC for LfD, we seek to solve the bilevel optimization problem 
\begin{align}
    \label{eq:prob-DT}
    \begin{split}
        \underset{\theta\in\Theta}{\text{min}} \quad & \mathcal{L}(\theta) = \sum_{t=0}^{\tau} L(y_t, g(x_t(\theta),u_t(\theta)))
    \end{split}
\end{align}
subject to the states $x_t(\theta) \in \mathbb{R}^n$ and controls $u_t(\theta) \in \mathbb{R}^m$ in~\eqref{eq:prob-DT} solving the \emph{discrete-time} optimal control problem
\begin{align}
	\label{eq:OCP-DT}
    \begin{split}
        \min_{x,u} \quad & h(x_T;\theta) + \sum_{t=0}^{T-1} \ell(x_t, u_t;\theta) \\
        \text{s.t.} \quad & x_{t+1} = f(x_t, u_t;\theta), \quad x_0 = x_0 \\
        & c_t(x_t, u_t; \theta) \leq 0, \quad \forall t\in\{0,\ldots,T-1\} \\
        & \bar{c}_t(x_t, u_t; \theta) = 0, \quad \forall t\in\{0,\ldots,T-1\} \\
        & c_T(x_T; \theta) \leq 0, \ \bar{c}_T(x_T; \theta) = 0
    \end{split}
\end{align}
and where in~\eqref{eq:prob-DT} we are given noisy partial measurements
\begin{equation}
    \label{eq:measurements-DT}
    y_t = g(x_t(\theta^*), u_t(\theta^*)) + w_t
\end{equation}
of the optimal states $x_t(\theta^*)$ and controls $u_t(\theta^*)$ solving~\eqref{eq:OCP-DT} with some unknown $\theta = \theta^*$.
Here, the parameters $\theta \in \Theta$, and functions defining the loss $L$, measurements $g$, costs $h,\ell$, dynamics $f$, and constraints $c_t, \bar{c}_t$ are defined as before, but $w_t : [0, \tau] \rightarrow \mathbb{R}^q$ is a (potentially stochastic) discrete-time noise process.
ZORMS-LfD in Algorithm~\ref{alg:ZO-RMS} can then be used to solve the optimization problem in~\eqref{eq:prob-DT} but the discrete-time optimal control problem~\eqref{eq:OCP-DT} is solved twice at each iteration in the oracle to evaluate the loss in~\eqref{eq:prob-DT} (instead of solving the continuous-time optimal control problem in~\eqref{eq:OCP-CT} twice at each iteration in the oracle to evaluate the loss in~\eqref{eq:prob-CT}).
Furthermore, the complexity bounds for ZORMS-LfD also hold for discrete-time IOC, with the loss $\mathcal{L}$ being that in~\eqref{eq:prob-DT}.

\section{Experimental Setup}\label{sec:experiments}

We evaluate the performance of ZORMS-LfD and competing methods in  both discrete and continuous time.\footnote{\url{https://github.com/olivi-dry/ZORMS-LfD}.}

\subsection{Continuous-Time Benchmark Problems and Algorithms}

To evaluate ZORMS-LfD for continuous-time IOC without constraints $c_t$ and $\bar{c}_t$, we use the four unconstrained benchmark problems detailed in~\cite{CPDP}: Cartpole, (2-link) Robot Arm, (6-DoF) Quadrotor, and (6-DoF) Rocket Landing.
Since no benchmark problems exist for continuous-times IOC with constraints, we created them by modifying the unconstrained benchmark problems to include constraints that match those in the constrained discrete-time benchmark problems detailed in~\cite{SafePDP}.
Where necessary, we constructed symmetric matrix parameters $\theta \in \Theta \subseteq \mathbb{S}^p$ using diagonal matrices with the parameter vector on  the diagonal.
The measurements in all problems are the position components of the state (i.e., rates are not measured, nor are controls) and have no added noise.
Table~\ref{table:envs} summarizes the problems.

On the unconstrained problems, we compare ZORMS-LfD against CPDP~\cite{CPDP}, which implements a first-order method with analytical gradients.
Since no dedicated algorithms exist for continuous-time IOC with constraints, we
compare against the Nelder-Mead algorithm, a commonly used derivative-free
optimization method{\color{black}, as well as
CMA-ES~\cite{hansen2023cmaevolutionstrategytutorial}, another stochastic
derivative-free optimization method}.
To ensure fair comparison, all algorithms use the same step size $\alpha$ and initialization $\hat{\theta}_0$, and the squared Euclidean norm as the loss.

\begin{table}[!t]
    \caption{Summary of Benchmark IOC Problems}\label{table:envs}
    \centering
    \begin{tabular}{|r|c|c|c|c|c|c|}
        \hline
	    Environment & $n$ & $m$ & \begin{tabular}[c]{@{}c@{}}$p$\\ (uncons.)\end{tabular} & \begin{tabular}[c]{@{}c@{}}$p$\\ (cons.)\end{tabular} & $\tau$ & \begin{tabular}[c]{@{}c@{}}$T$\\ (s)\end{tabular} \\
        \hline
        Cartpole & 4 & 1 & 7 & 9 & 10 & 3 \\
        Robot Arm & 4 & 2 & 8 & 10 & 10 & 3.5 \\
        Quadrotor & 13 & 4 & 9 & 11 & 10 & 5 \\
        Rocket & 13 & 3 & 10 & 12 & 10 & 4 \\
        \hline
    \end{tabular}
\end{table}

\subsection{Discrete-Time Benchmark Problems and Algorithms}

To evaluate ZORMS-LfD for discrete-time IOC without and with constraints, we use the same four benchmark problems as in continuous time but in their discrete-time forms detailed in~\cite{PDP_framework} and~\cite{SafePDP}.
We constructed matrix parameters and the measurements in the same manner as in the continuous-time benchmark problems.
The dimensions of the problems are as summarized in Table~\ref{table:envs}.


We compare ZORMS-LfD against existing state-of-the-art gradient-based methods.
Specifically, on the unconstrained problems we compare against PDP~\cite{PDP_framework} and IDOC~\cite{COC_problems}, whilst on the constrained problems we compare against SafePDP~\cite{SafePDP} and IDOC~\cite{COC_problems}.
For SafePDP, we compare against both strategy (a), where the gradient is obtained directly from the constrained optimal control problem, and strategy (b), where a log-barrier problem is established.
All algorithms use the same step size $\alpha$, initialization $\hat{\theta}_0$, and squared Euclidean norm loss.

\subsection{Learning from Human Demonstrations}

We also apply ZORMS-LfD to learning a manipulation task {\color{black} for a 7 DoF arm from (real) motion-capture data of a human}.
We use the ``Motion \#1278'' dataset of a wave from the KIT Whole-Body Human Motion Dataset~\cite{KIT-Data}.
We assume that the control scheme is constructed in the joint-space.
As such, the motion will be described by the trajectory of the joint angles.
The human arm kinematic model (as defined by~\cite{KIT-Data}) has seven revolute joints, with three occurring at the shoulder, two at the elbow, and two at the wrist.
The state is the collection of the joint angles $x={(\phi_1, \phi_2,\ldots,\phi_7)}^\top \in\mathbb{R}^7$, and the control is the angular velocity input for each joint $u\in\mathbb{R}^7$.
We modeled a linearized system, and seek to learn the cost matrices of a linear-quadratic tracking controller, thus $\theta$ are the cost matrices from $\Theta \subseteq \mathbb{S}^{21}$.
{\color{black} We learn the tracking controller from the \texttt{wave\_right01} sequence and test it tracking the \texttt{wave\_right02} sequence.
To illustrate the generality of the learned controller, we implemented it for the 5 DoF arm of a Nao robot in Webots\footnote{https://cyberbotics.com} (simply omitting the weights for the joints not present) and used it to track a wave demonstration.
}

\section{Experimental Results}\label{sec:results}

We report the losses and compute times as results.

\subsection{Continuous-Time Benchmark Problems}
\begin{table}[!t]
    \caption{\color{black}Unconstrained Continuous-Time Benchmark Results}\label{table:uncons-results-ct}
    \centering
    \begin{tabular}{|c c||c|c|}
        \hline
        \multicolumn{2}{|c||}{Environment} & Min. $\mathcal{L}(\hat{\theta})$ & Avg. $N$ \\
        \hline
        \multicolumn{1}{|c|}{\multirow{2}{*}{Cartpole}} & ZORMS-LfD & 0.000033 & \textbf{41} \\ \cline{2-4}
        \multicolumn{1}{|c|}{} & CPDP & \textbf{0.000013} & 45 \\
        \hline
        \multicolumn{1}{|c|}{\multirow{2}{*}{\begin{tabular}[c]{@{}c@{}}Robot\\ Arm\end{tabular}}} & ZORMS-LfD & \textbf{0.0019} & \textbf{43} \\ \cline{2-4}
        \multicolumn{1}{|c|}{} & CPDP & 0.022 & 44 \\
        \hline
        \multicolumn{1}{|c|}{\multirow{2}{*}{Quadrotor}} & ZORMS-LfD & \textbf{0.0034} & \textbf{45} \\ \cline{2-4}
        \multicolumn{1}{|c|}{} & CPDP & 0.010 & 46 \\
        \hline
        \multicolumn{1}{|c|}{\multirow{2}{*}{Rocket}} & ZORMS-LfD & \textbf{0.00091} & 43 \\ \cline{2-4}
        \multicolumn{1}{|c|}{} & CPDP & 0.0073 & \textbf{42} \\
        \hline
    \end{tabular}
\end{table}

\begin{figure}[!t]
    \centering
    \subfloat[]{\includegraphics[width=0.48\columnwidth]{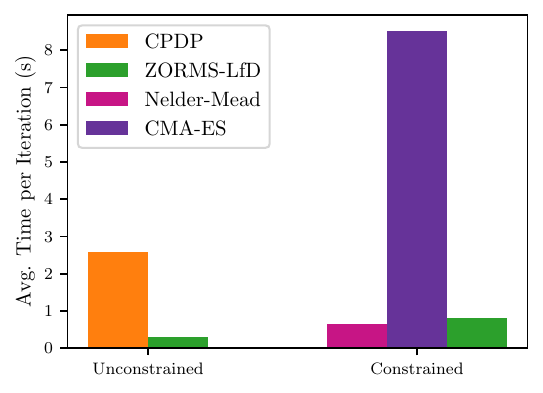}\label{fig:timing-ct}}
    \subfloat[]{\includegraphics[width=0.48\columnwidth]{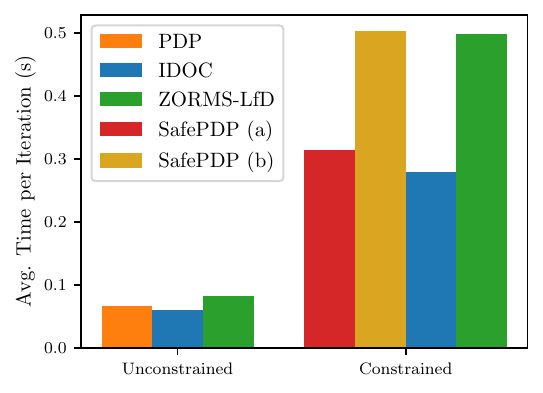}\label{fig:timing-dt}}
    \caption{Average compute times on all benchmark problems: (a) Unconstrained and Constrained in Continuous-time, (b) Unconstrained and Constrained Discrete-time.
    \vspace{-1.5em}
    }\label{fig:timing}
\end{figure}

\begin{figure*}[t!]
    \centering
    \subfloat[Cartpole]{\includegraphics[width=0.24\textwidth]{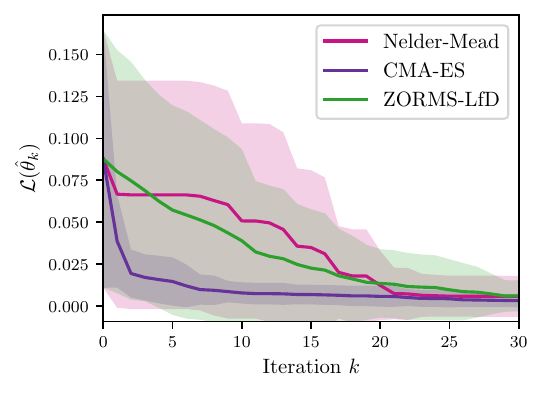}\label{fig:cons-results-cartpole-ct}}
    \hfil
    \subfloat[Robot Arm]{\includegraphics[width=0.24\textwidth]{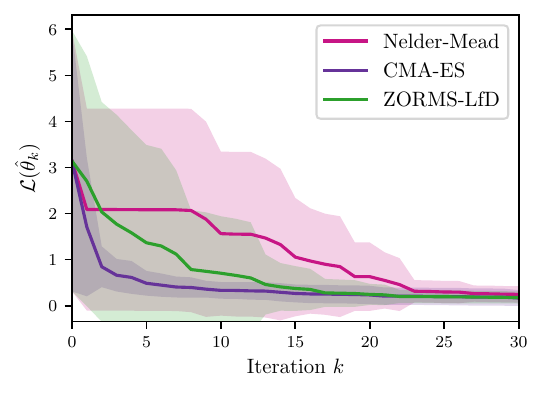}\label{fig:cons-results-arm-ct}}
    \hfil
    \subfloat[Quadrotor]{\includegraphics[width=0.24\textwidth]{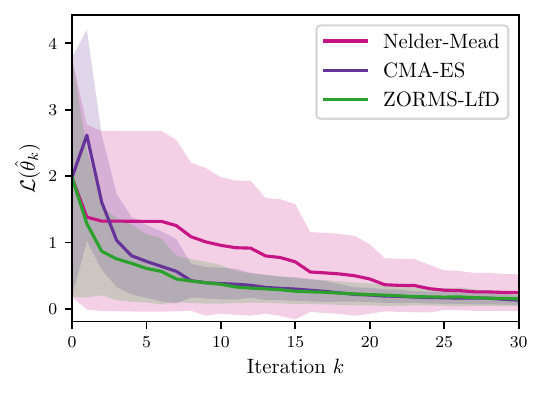}\label{fig:cons-results-quadrotor-ct}}
    \hfil
    \subfloat[Rocket]{\includegraphics[width=0.24\textwidth]{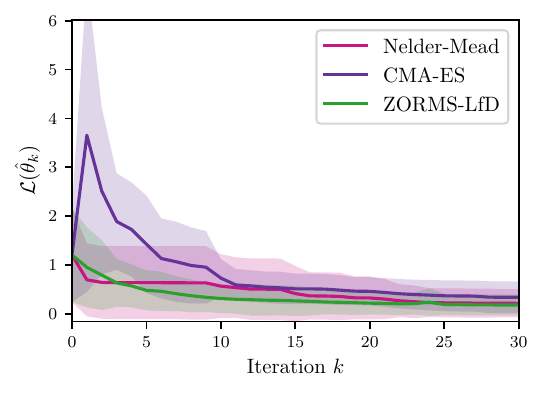}\label{fig:cons-results-rocket-ct}}
    \caption{Constrained Continuous-Time Benchmark Problem Losses. Solid lines are mean values and shaded area is one standard deviation over 25 trials.
    \vspace{-1.5em}
    }\label{fig:cons-results-ct}
\end{figure*}

{\color{black}Table~\ref{table:uncons-results-ct} reports the minimum value of the loss after 50 iterations, and the average number of iterations to converge within 5\% of this minimum loss on 25 trials of the continuous-time unconstrained benchmark problems.
Figure~\ref{fig:cons-results-ct} reports the losses after each iteration from 25 trials of the continuous-time constrained benchmark problems.
The average compute times per iteration are reported in Fig.~\ref{fig:timing-ct}.}


{\color{black}
From Table~\ref{table:uncons-results-ct}, we see that ZORMS-LfD has similar loss and convergence behavior to CPDP.
This result is surprising since CPDP is a first-order method, which are typically expected to outperform zeroth-order methods due to the analytical gradients.
Further, from Fig.~\ref{fig:timing-ct} we see that each iteration of ZORMS-LfD is over $80\%$ faster than CPDP, despite CPDP involving the solution of one optimal control problem~\eqref{eq:OCP-CT} at each iteration whilst ZORMS-LfD involves the solution of two.}
However, the gradient computation in CPDP takes significantly longer than solving an optimal control problem, in part because it requires integrating two (coupled) sets of ordinary differential equations~\cite{CPDP}.
Furthermore, unlike ZORMS-LfD's ability to solve the two optimal controls problems in parallel (although not exploited here), the solution of the two ordinary differential equations in CPDP cannot be parallelized.

%

From Fig.~\ref{fig:cons-results-ct}, we observe that{\color{black}, for the \emph{constrained} problems,} ZORMS-LfD reliably matches or exceeds the performance of the Nelder-Mead algorithm for the same number of iterations.
The compute time per iteration is also similar (cf.\ Fig.~\ref{fig:timing-ct}).
However, ZORMS-LfD has convergence guarantees established in Section~\ref{sec:zorms}, whilst Nelder-Mead lacks similar performance guarantees.
{\color{black} On the other hand, the CMA-ES algorithm outperforms ZORMS-LfD on Cartpole and Robot Arm, but performs significantly worse on the more complex Quadrotor and Rocket problems.
Additionally, the compute time per iteration is about 8 times slower, due to many more loss functions evaluations every iteration.}

\subsection{Discrete-Time Benchmark Problems}


\begin{table}[!t]
    \caption{\color{black}Unconstrained Discrete-Time Benchmark Results}\label{table:uncons-results-dt}
    \centering
    \begin{tabular}{|c c||c|c|}
        \hline
        \multicolumn{2}{|c||}{Environment} & Min. $\mathcal{L}(\hat{\theta})$ & Avg. $N$ \\
        \hline
        \multicolumn{1}{|c|}{\multirow{2}{*}{Cartpole}} & ZORMS-LfD & \textbf{0.016} & \textbf{89} \\ \cline{2-4}
        \multicolumn{1}{|c|}{} & PDP \& IDOC & 0.017 & 91 \\
        \hline
        \multicolumn{1}{|c|}{\multirow{2}{*}{\begin{tabular}[c]{@{}c@{}}Robot\\ Arm\end{tabular}}} & ZORMS-LfD & \textbf{0.15} & \textbf{90} \\ \cline{2-4}
        \multicolumn{1}{|c|}{} & PDP \& IDOC & 0.65 & 93 \\
        \hline
        \multicolumn{1}{|c|}{\multirow{2}{*}{Quadrotor}} & ZORMS-LfD & 0.44 & \textbf{92} \\ \cline{2-4}
        \multicolumn{1}{|c|}{} & PDP \& IDOC & \textbf{0.34} & 93 \\ \cline{2-4}
        \hline
        \multicolumn{1}{|c|}{\multirow{2}{*}{Rocket}} & ZORMS-LfD & \textbf{0.99} & \textbf{83} \\ \cline{2-4}
        \multicolumn{1}{|c|}{} & PDP \& IDOC & 1.54 & 85 \\
        \hline
    \end{tabular}
\end{table}

\begin{figure*}[!t]
    \centering
    \subfloat[Cartpole]{\includegraphics[width=0.24\textwidth]{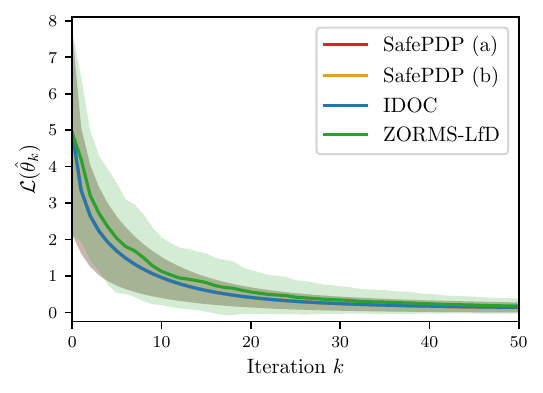}\label{fig:cons-results-cartpole-dt}}
    \hfil
    \subfloat[Robot Arm]{\includegraphics[width=0.24\textwidth]{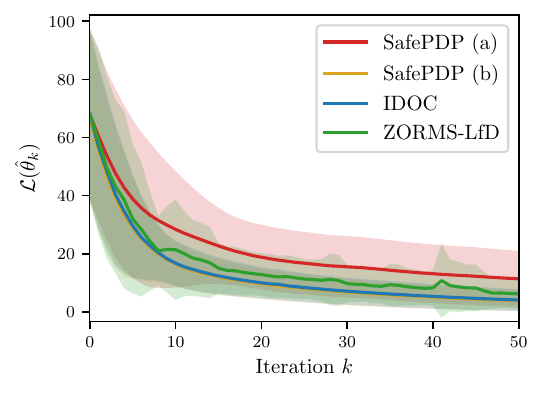}\label{fig:cons-results-arm-dt}}
    \hfil
    \subfloat[Quadrotor]{\includegraphics[width=0.24\textwidth]{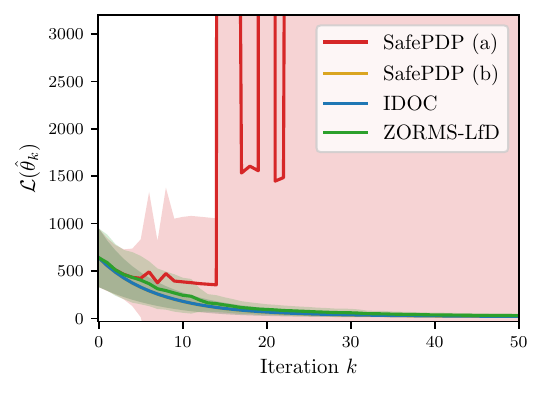}\label{fig:cons-results-quadrotor-dt}}
    \hfil
    \subfloat[Rocket]{\includegraphics[width=0.24\textwidth]{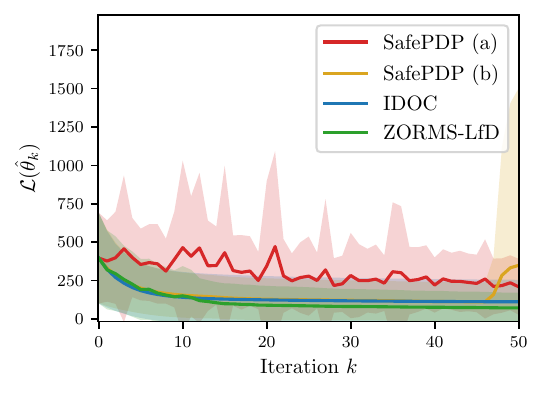}\label{fig:cons-results-rocket-dt}}
    \caption{Constrained Discrete-time Benchmark Problem Losses. Solid lines are mean values and shaded area is one standard deviation over 25 trials.
    \vspace{-1.5em}
    }\label{fig:cons-results-dt}
\end{figure*}


{\color{black}Table~\ref{table:uncons-results-dt} reports the minimum value of the loss, and the average number of iterations to converge within 5\% of this minimum loss on 25 trials of the discrete-time unconstrained benchmark problems.
Figure~\ref{fig:cons-results-dt} reports the losses after each iteration from 25 trials of the discrete-time constrained benchmark problems.
The average compute times per iteration are reported in Fig.~\ref{fig:timing-dt}.}


Overall, the losses and compute times of ZORMS-LfD is again surprisingly similar to the other methods, all of which employ first-order optimization and differ only in how they compute gradients of the loss.
Most notably, ZORMS-LfD offers more consistent performance in the constrained discrete-time problems (cf.\ Fig.~\ref{fig:cons-results-dt}) where the presence of (parameterized) constraints means that the loss $\mathcal{L}$ is not (globally) smooth.
Thus, in some cases the first-order methods that (erroneously) assume that the loss is continuously differentiable, fail to compute a sensible gradient.
These failures are most evident in Fig.~\ref{fig:cons-results-rocket-dt} for Rocket where both IDOC and SafePDP (b) fail to compute gradients in 5 of the 25 trials (though not necessarily in the same trials).



\subsection{Resilience to Measurement Noise}



\begin{figure}[!t]
    \centering
    \includegraphics[width=0.4\textwidth]{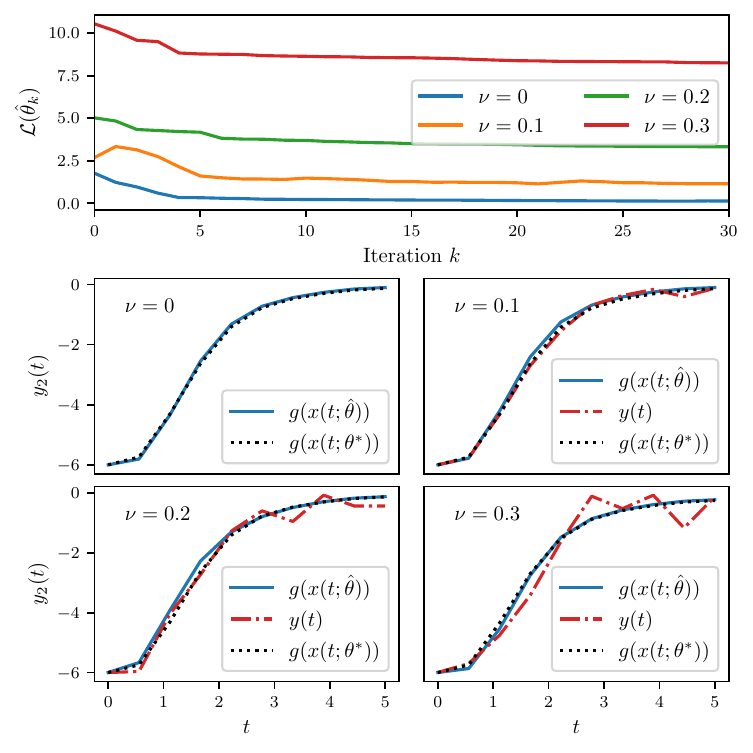}
    \caption{Results for the noise study on the constrained quadrotor system. In the top plot, the solid lines represent the mean value of the loss for over 5 trials for different amounts of added noise $\nu$. The bottom plots are the measured trajectories using the determined parameters, compared with the noiseless reference measurement, and the reference with added noise.
    \vspace{-1.5em}
    }\label{fig:noise}
\end{figure}

To illustrate the effect of noise, we use the constrained continuous-time Quadrotor problem and add Gaussian noise with increasing variance to the measurements.
Specifically, we consider measurements of the form $y(t) = g(x(t), u(t)) + \nu w, \quad w\sim \mathcal{N}(0,1)$ with $\nu\in\{0, 0.1, 0.2, 0.3\}$.
The convergence of the loss with noisy measurements is shown in Fig.~\ref{fig:noise}.
When more noise is added to the reference samples, the minimum loss is much greater.
We also observe that in the comparison of the determined trajectories with the reference measurements, the trajectories generated using the determined parameters are very similar to the noiseless reference trajectory.
This indicates that ZORMS-LfD is able to determine a sufficient estimation of the true parameters despite noisy partial measurements.



\subsection{Learning from Human Demonstrations}

\begin{figure}[t!]
    \centering
    \includegraphics[width=0.286\textwidth]{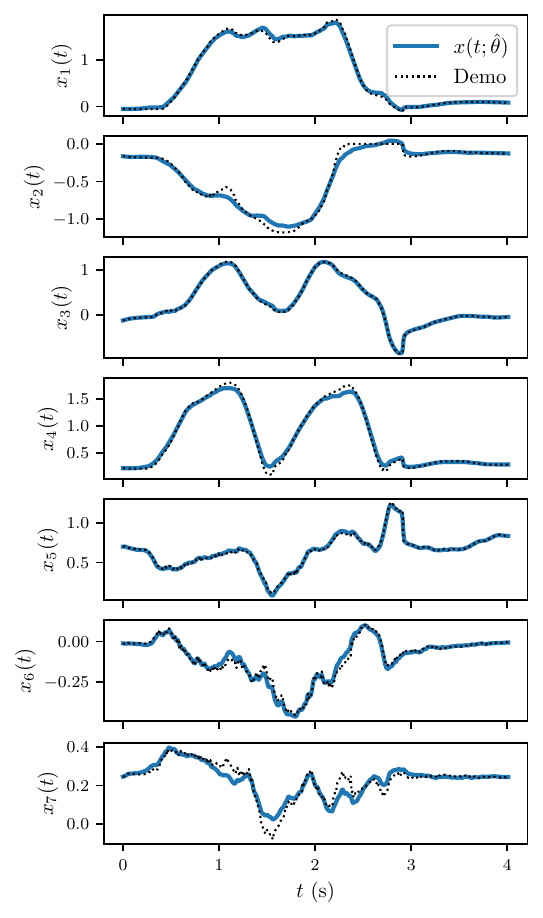}
    \caption{The state trajectories for learning from human demonstration with the learned (tracking) controller applied to a test dataset.
    The human demonstration (test) data is represented by the dotted line.
    \vspace{-1.5em}
    }\label{fig:real-data-state}
\end{figure}

Results on the human demonstration (test) data are shown in Fig.~\ref{fig:real-data-state} {\color{black} (and the video attachment)}. 
ZORMS-LfD reduced the loss significantly in only ten iterations.
Further, we see that the state trajectory using the learned controller closely follows the reference (test) trajectory.
{\color{black}The dimension of the parameter space in this problem is much larger than the benchmark problems, thus we observe that ZORMS-LfD is still able to perform in large-scale matrix parameter spaces.}

\section{Conclusion}\label{sec:conclusion}

We introduce ZORMS-LfD for learning optimal control problems from noisy partial measurements of demonstrations.
ZORMS-LfD is applicable to both constrained and unconstrained problems, and both continuous- and discrete-time systems, given the underlying (forward) optimal control problem can be solved.
ZORMS-LfD avoids the challenge and assumptions associated with finding gradients of the learning loss by using a zeroth-order random matrix oracle to construct a smooth version of the loss and its directional derivatives in randomly chosen directions.
Additionally, ZORMS-LfD is shown to enjoy complexity bounds, which provide insight into the selection of its hyperparameters.
The performance of ZORMS-LfD on benchmarks shows that its convergence and compute times are similar to those of existing state-of-the-art first-order methods despite being a zeroth-order method.
ZORMS-LfD thus establishes a new state-of-the-art in solving continuous-time inverse optimal control when the underlying optimal control problem is constrained, and a new benchmark in reliably solving inverse optimal control problems in both discrete and continuous time when the loss function is nonsmooth.

\bibliography{references}

\begin{thebibliography}{10}

\bibitem{locomotion_mombaur}
K.~Mombaur, A.~Truong, and J.-P. Laumond, ``From human to humanoid
  locomotion-an inverse optimal control approach,'' {\em Auton. Robots},
  vol.~28, pp.~369--383, 04 2010.

\bibitem{Dana-Kulic-Motion}
J.~F.-S. Lin, V.~Bonnet, A.~M. Panchea, N.~Ramdani, G.~Venture, and
  D.~Kuli{\'c}, ``Human motion segmentation using cost weights recovered from
  inverse optimal control,'' in {\em 2016 IEEE-RAS 16th International
  Conference on Humanoid Robots (Humanoids)}, pp.~1107--1113, 2016.

\bibitem{jumping}
K.~Westermann, J.~Lin, and D.~Kulic, ``Inverse optimal control with
  time-varying objectives: application to human jumping movement analysis,''
  {\em Scientific Reports}, vol.~10, p.~11174, 07 2020.

\bibitem{rebula2019robustness}
J.~R. Rebula, S.~Schaal, J.~Finley, and L.~Righetti, ``A robustness analysis of
  inverse optimal control of bipedal walking,'' {\em IEEE Robotics and
  Automation Letters}, vol.~4, no.~4, pp.~4531--4538, 2019.

\bibitem{neumeyer2021general}
C.~Neumeyer, F.~A. Oliehoek, and D.~M. Gavrila, ``General-sum multi-agent
  continuous inverse optimal control,'' {\em IEEE Robotics and Automation
  Letters}, vol.~6, no.~2, pp.~3429--3436, 2021.

\bibitem{Abbeel2010}
P.~Abbeel, A.~Coates, and A.~Y. Ng, ``Autonomous helicopter aerobatics through
  apprenticeship learning,'' {\em The International Journal of Robotics
  Research}, vol.~29, no.~13, pp.~1608--1639, 2010.

\bibitem{DiffMPC}
B.~Amos, I.~D.~J. Rodriguez, J.~Sacks, B.~Boots, and J.~Z. Kolter,
  ``Differentiable mpc for end-to-end planning and control,'' in {\em
  Proceedings of the 32nd International Conference on Neural Information
  Processing Systems}, NIPS'18, (Red Hook, NY, USA), p.~8299–8310, Curran
  Associates Inc., 2018.

\bibitem{PDP_framework}
W.~Jin, Z.~Wang, Z.~Yang, and S.~Mou, ``Pontryagin differentiable programming:
  An end-to-end learning and control framework,'' in {\em Advances in Neural
  Information Processing Systems}, vol.~33, pp.~7979--7992, Curran Associates,
  Inc., 2020.

\bibitem{SafePDP}
W.~Jin, S.~Mou, and G.~J. Pappas, ``Safe pontryagin differentiable
  programming,'' in {\em Advances in Neural Information Processing Systems},
  vol.~34, pp.~16034--16050, Curran Associates, Inc., 2021.

\bibitem{CPDP}
W.~Jin, T.~D. Murphey, D.~Kuli{\'c}, N.~Ezer, and S.~Mou, ``Learning from
  sparse demonstrations,'' {\em IEEE Transactions on Robotics}, vol.~39, no.~1,
  pp.~645--664, 2023.

\bibitem{COC_problems}
M.~Xu, T.~L. Molloy, and S.~Gould, ``Revisiting implicit differentiation for
  learning problems in optimal control,'' in {\em Advances in Neural
  Information Processing Systems}, vol.~36, pp.~60060--60076, Curran
  Associates, Inc., 2023.

\bibitem{Molloy2022}
T.~L. Molloy, J.~I. Charaja, S.~Hohmann, and T.~Perez, {\em Inverse optimal
  control and inverse noncooperative dynamic game theory}.
\newblock Springer, 2022.

\bibitem{imputing}
A.~Keshavarz, Y.~Wang, and S.~Boyd, ``Imputing a convex objective function,''
  in {\em 2011 IEEE International Symposium on Intelligent Control},
  pp.~613--619, 2011.

\bibitem{inverseKKT}
P.~Englert, N.~A. Vien, and M.~Toussaint, ``Inverse kkt: Learning cost
  functions of manipulation tasks from demonstrations,'' {\em The International
  Journal of Robotics Research}, vol.~36, no.~13-14, pp.~1474--1488, 2017.

\bibitem{convex_loco}
A.-S. Puydupin-Jamin, M.~Johnson, and T.~Bretl, ``A convex approach to inverse
  optimal control and its application to modeling human locomotion,'' in {\em
  2012 IEEE International Conference on Robotics and Automation}, pp.~531--536,
  2012.

\bibitem{pontry}
T.~L. Molloy, J.~J. Ford, and T.~Perez, ``Finite-horizon inverse optimal
  control for discrete-time nonlinear systems,'' {\em Automatica}, vol.~87,
  pp.~442--446, 2018.

\bibitem{cont_time}
M.~Johnson, N.~Aghasadeghi, and T.~Bretl, ``Inverse optimal control for
  deterministic continuous-time nonlinear systems,'' in {\em 52nd IEEE
  Conference on Decision and Control}, pp.~2906--2913, 2013.

\bibitem{diff_flat}
N.~Aghasadeghi and T.~Bretl, ``Inverse optimal control for differentially flat
  systems with application to locomotion modeling,'' in {\em 2014 IEEE
  International Conference on Robotics and Automation (ICRA)}, pp.~6018--6025,
  2014.

\bibitem{IOC_polynom}
E.~Pauwels, D.~Henrion, and J.-B. Lasserre, ``Inverse optimal control with
  polynomial optimization,'' in {\em 53rd IEEE Conference on Decision and
  Control}, pp.~5581--5586, 2014.

\bibitem{robust_approach}
J.~Thai and A.~M. Bayen, ``Imputing a variational inequality function or a
  convex objective function: A robust approach,'' {\em Journal of Mathematical
  Analysis and Applications}, vol.~457, no.~2, pp.~1675--1695, 2018.
\newblock Special Issue on Convex Analysis and Optimization: New Trends in
  Theory and Applications.

\bibitem{suh2022bundled}
H.~J.~T. Suh, T.~Pang, and R.~Tedrake, ``Bundled gradients through contact via
  randomized smoothing,'' {\em IEEE Robotics and Automation Letters}, vol.~7,
  no.~2, pp.~4000--4007, 2022.

\bibitem{suh2022differentiable}
H.~J. Suh, M.~Simchowitz, K.~Zhang, and R.~Tedrake, ``Do differentiable
  simulators give better policy gradients?,'' in {\em International Conference
  on Machine Learning}, pp.~20668--20696, PMLR, 2022.

\bibitem{le2024leveraging}
Q.~Le~Lidec, F.~Schramm, L.~Montaut, C.~Schmid, I.~Laptev, and J.~Carpentier,
  ``Leveraging randomized smoothing for optimal control of nonsmooth dynamical
  systems,'' {\em Nonlinear Analysis: Hybrid Systems}, vol.~52, p.~101468,
  2024.

\bibitem{zorms}
A.~I. Maass, C.~Manzie, I.~Shames, and H.~Nakada, ``Zeroth-order optimization
  on subsets of symmetric matrices with application to {MPC} tuning,'' {\em
  IEEE Transactions on Control Systems Technology}, vol.~30, no.~4,
  pp.~1654--1667, 2022.

\bibitem{Anderson_Guionnet_Zeitouni_2009}
G.~W. Anderson, A.~Guionnet, and O.~Zeitouni, {\em An Introduction to Random
  Matrices}.
\newblock Cambridge Studies in Advanced Mathematics, Cambridge University
  Press, 2009.

\bibitem{vec_oracle}
Y.~Nesterov and V.~G. Spokoiny, ``Random gradient-free minimization of convex
  functions,'' {\em Foundations of Computational Mathematics}, vol.~17, pp.~527
  -- 566, 2015.

\bibitem{moments}
T.~Yumiko~Hayakawa and Kikuchi, ``The moments of a function of traces of a
  matrix with a multivariate symmetric normal distribution,'' {\em South
  African Statistical Journal}, vol.~13, no.~1, pp.~71--82, 1979.

\bibitem{ghadimi2013}
S.~Ghadimi, G.~Lan, and H.~Zhang, ``Mini-batch stochastic approximation methods
  for nonconvex stochastic composite optimization,'' {\em Math. Program.},
  vol.~155, p.~267–305, Jan. 2016.

\bibitem{Liu2018}
S.~Liu, X.~Li, P.-Y. Chen, J.~Haupt, and L.~Amini, ``Zeroth-order stochastic
  projected gradient descent for nonconvex optimization,'' in {\em 2018 IEEE
  Global Conference on Signal and Information Processing (GlobalSIP)},
  pp.~1179--1183, 2018.

\bibitem{bunin2017lipschitzconstantsexperimentaloptimization}
G.~A. Bunin and G.~François, ``Lipschitz constants in experimental
  optimization,'' 2017.

\bibitem{hansen2023cmaevolutionstrategytutorial}
N.~Hansen, ``The cma evolution strategy: A tutorial,'' 2023.

\bibitem{KIT-Data}
C.~Mandery, O.~Terlemez, M.~Do, N.~Vahrenkamp, and T.~Asfour, ``Unifying
  representations and large-scale whole-body motion databases for studying
  human motion,'' {\em IEEE Transactions on Robotics}, vol.~32, no.~4,
  pp.~796--809, 2016.

\end{thebibliography}
\bibliographystyle{ieeetr}


\end{document}